\documentclass[nohyperref]{article}

\usepackage{microtype}
\usepackage{graphicx}
\usepackage{subcaption}
\usepackage{booktabs} %

\usepackage{hyperref}

\graphicspath{{img/}}

\newcommand{\legend}[2]{
	\begin{figure*}[tb]
		\centering
	    \includegraphics[width=0.4\textwidth]{#1}
		\label{fig:#2}
		\vspace{-0.24cm}
	\end{figure*}
}

\newcommand{\picTwo}[6]{
	\begin{figure*}[tb]
		\centering
		\begin{subfigure}[c]{0.49\textwidth}
			\centering		
			\includegraphics[width=\textwidth]{#1}
			\subcaption{#4}	
			\label{fig:#3-0}	
		\end{subfigure}
		\begin{subfigure}[c]{0.49\textwidth}	
			\centering	
			\includegraphics[width=\textwidth]{#2}
			\subcaption{#5}
			\label{fig:#3-1}	
		\end{subfigure}
		\caption{#6}
		\label{fig:#3}
	\end{figure*}
}

\newcommand{\picFour}[9]{
	\begin{figure*}[tb]
		\centering
		\begin{subfigure}[c]{0.49\textwidth}
			\centering		
			\includegraphics[width=\textwidth]{#1}
			\subcaption{#5}	
		\end{subfigure}
		\begin{subfigure}[c]{0.49\textwidth}	
			\centering	
			\includegraphics[width=\textwidth]{#2}
			\subcaption{#6}		
		\end{subfigure}
		\begin{subfigure}[c]{0.49\textwidth}
			\centering		
			\includegraphics[width=\textwidth]{#3}
			\subcaption{#7}		
		\end{subfigure}
		\begin{subfigure}[c]{0.49\textwidth}	
			\centering	
			\includegraphics[width=\textwidth]{#4}
			\subcaption{#8}		
		\end{subfigure}
		\caption{#9}
		\label{fig:#1}
	\end{figure*}
}
\usepackage{xspace}
\newcommand{\acc}{$ACC$\xspace}

\newcommand{\kl}{$KL$\xspace}

\newcommand{\tblResults}{
\begin{table}[tb]
	\caption{ Quantitative comparison between hard and soft labels --
	A small arrow after the metrics indicates if lower or higher values are better. 
	The best results per metric and dataset are marked bold.
	}
	\resizebox{\linewidth}{!}{

		\begin{tabular}{l c c c c }
			\toprule
			Dataset & \multicolumn{2}{c}{Using Hard Labels} & \multicolumn{2}{c}{Using Soft Labels} \\

			\cmidrule(r){2-3} \cmidrule(r){4-5}
			
			& \acc ~$\uparrow$ & \kl ~$\downarrow$ & \acc ~$\uparrow$ & \kl ~$\downarrow$ \\
			
			\midrule
			
			Synthetic & 0.8247 $\pm$ 0.05002 & 0.4137 $\pm$ 0.0836 &  \textbf{0.9096 $\pm$ 0.0137} & \textbf{0.0394 $\pm$ 0.0009}\\
			Mice Bone & 0.6217 $\pm$ 0.04658 & 0.7884 $\pm$ 0.1089 & \textbf{0.6903 $\pm$ 0.0669} & \textbf{0.2280 $\pm$ 0.0227}\\
			\bottomrule

		\end{tabular}
	}
		
	\label{tbl:datasets}
\end{table}
}

\usepackage{comment}

\usepackage[accepted]{icml2022}

\usepackage{amsmath}
\usepackage{amssymb}
\usepackage{mathtools}
\usepackage{amsthm}

\usepackage[capitalize,noabbrev]{cleveref}

\theoremstyle{plain}

\theoremstyle{definition}

\theoremstyle{remark}

\usepackage[textsize=tiny]{todonotes}

\icmltitlerunning{Beyond Hard Labels: Investigating data label distributions}

\begin{document}

\twocolumn[
\icmltitle{Beyond Hard Labels: Investigating data label distributions}

\icmlsetsymbol{equal}{*}

\begin{icmlauthorlist}
\icmlauthor{Vasco Grossmann}{equal,mip}
\icmlauthor{Lars Schmarje}{equal,mip}
\icmlauthor{Reinhard Koch}{mip}
\end{icmlauthorlist}

\icmlaffiliation{mip}{Multimedia Information Processing Group, University of Kiel, Germany, cor}

\icmlcorrespondingauthor{Vasco Grossmann}{vgr@informatik.uni-kiel.de}
\icmlcorrespondingauthor{Lars Schmarje}{las@informatik.uni-kiel.de}

\icmlkeywords{Machine Learning, ICML}

\vskip 0.3in
]

\printAffiliationsAndNotice{\icmlEqualContribution} %

\begin{abstract}
High-quality data is a key aspect of modern machine learning. 
However, labels generated by humans suffer from issues like label noise and class ambiguities.
We raise the question of whether hard labels are sufficient to represent the underlying ground truth distribution in the presence of these inherent imprecision.
Therefore, we compare the disparity of learning with hard and soft labels quantitatively and qualitatively for a synthetic and a real-world dataset.
We show that the application of soft labels leads to improved performance and yields a more regular structure of the internal feature space.
\end{abstract}

\section{Motivation}
Modern machine learning relies on high-quality data, but even large and manually cleaned datasets like ImageNet contain errors and uncertainties \cite{relabelImagenet,Northcutt2021pervasiveErrors}.
In a model-centric view \cite{santarossa2022medregnet,Damm2021Sofia}, we could try to increase the robustness of models to overcome such issues.
However, in this work, we take a data-centric perspective and investigate the possibilities of improving data-quality \cite{Marcu2021datacentricMyths,schmarje2021datacentric}. 
In machine learning, we compare model predictions with ground-truth labels to measure the model performance and inherently also the data-quality.
As ground-truth, we often use class labels created by humans and indirectly expect them to be perfect for the evaluation.

This approach has two major shortcomings: errors and uncertainties.
Firstly, when humans create labels, errors and mistakes are unavoidable.
Even extensive cleaning does not remove all of these issues as recent works showed \cite{are_we_done, Northcutt2021pervasiveErrors}.
Thus, we still have to expect incorrect labels in our ground truth data.
We call these errors \emph{noise} \cite{divide-mix} and they lead to partially false evaluation results.
Secondly, the human perception of image content can vary. 
Recent work \cite{cifar10h,Wei2021cifar10n} showed that human provide different labels e.g., even for classifications of cats and dogs.
This uncertainty can arise from different factors like subjective interpretations \cite{cancergrading}, imperfect image qualities \cite{cifar10h} or arbitrary class distinctions \cite{are_we_done}.
We call this issue of data uncertainty \emph{ambiguity}.

In the literature, we find more robust methods \cite{liu2020elr} or improved datasets \cite{are_we_done,Northcutt2021pervasiveErrors} to resolve the issues of noise and ambiguity. 
Most approaches share the common assumption that one hard label per image as ground truth is sufficient  to capture the image information.
In theory, this approach can remove all noise in our labels with sufficient effort, but the literature also shows that ambiguity is present in many real world datasets \cite{cifar10h,Wei2021cifar10n,schmarje2022benchmark,benthic_uncertainty,dc3,foc}.
Thus, we must use a consensus process or a majority vote if we want to collapse the annotations to one hard label.
We raise the question whether such a collapse can represent the inherent ambiguity in the images.

We investigate the label distributions of one synthetic and one real-world dataset to answer this question.
The focus lies on the comparison between hard and soft labels as input for the model.
We evaluate the impact of these two representations quantitatively and qualitatively for estimating the label distribution.
We discuss the implications of our findings for future data-quality estimations.

\picFour{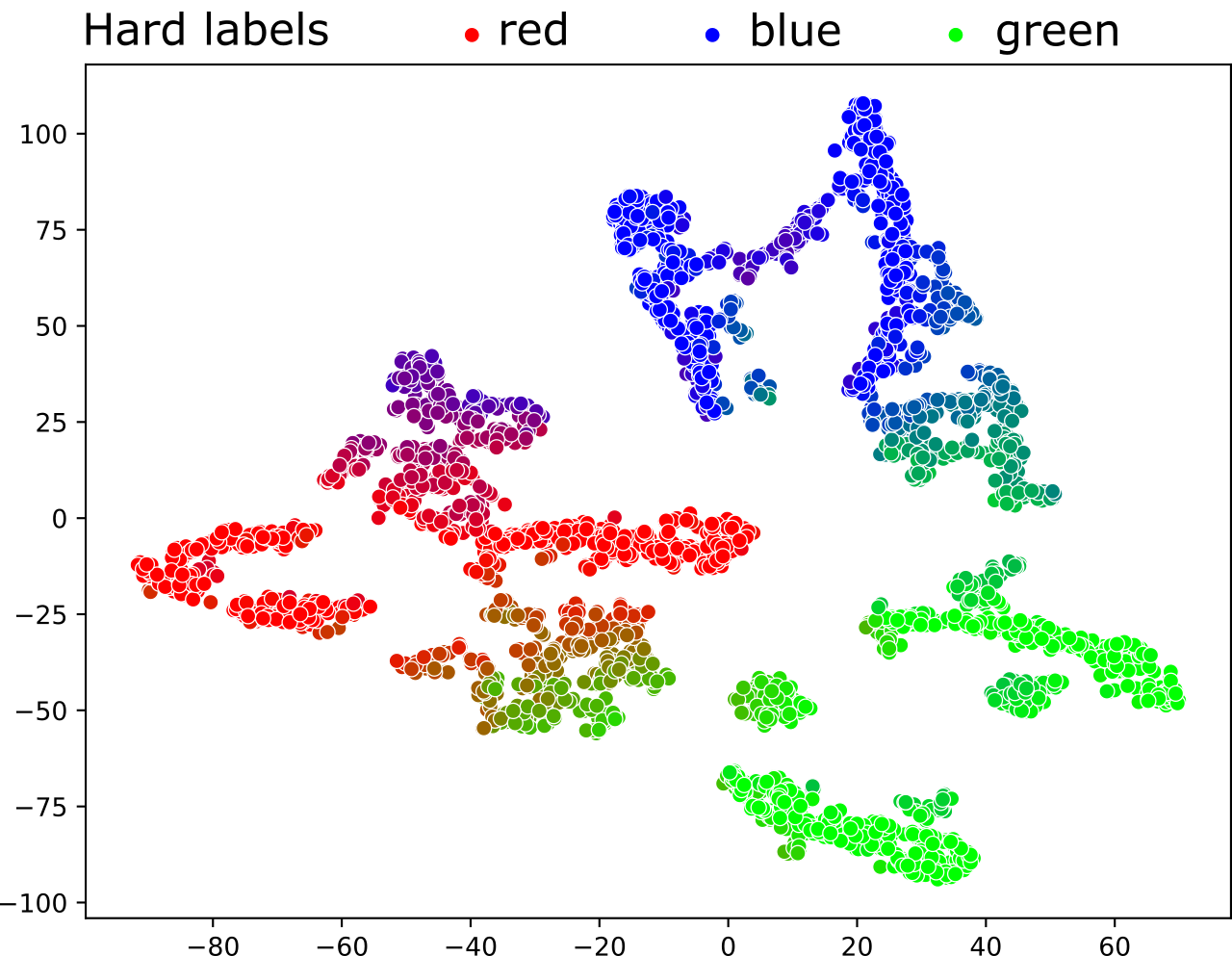}{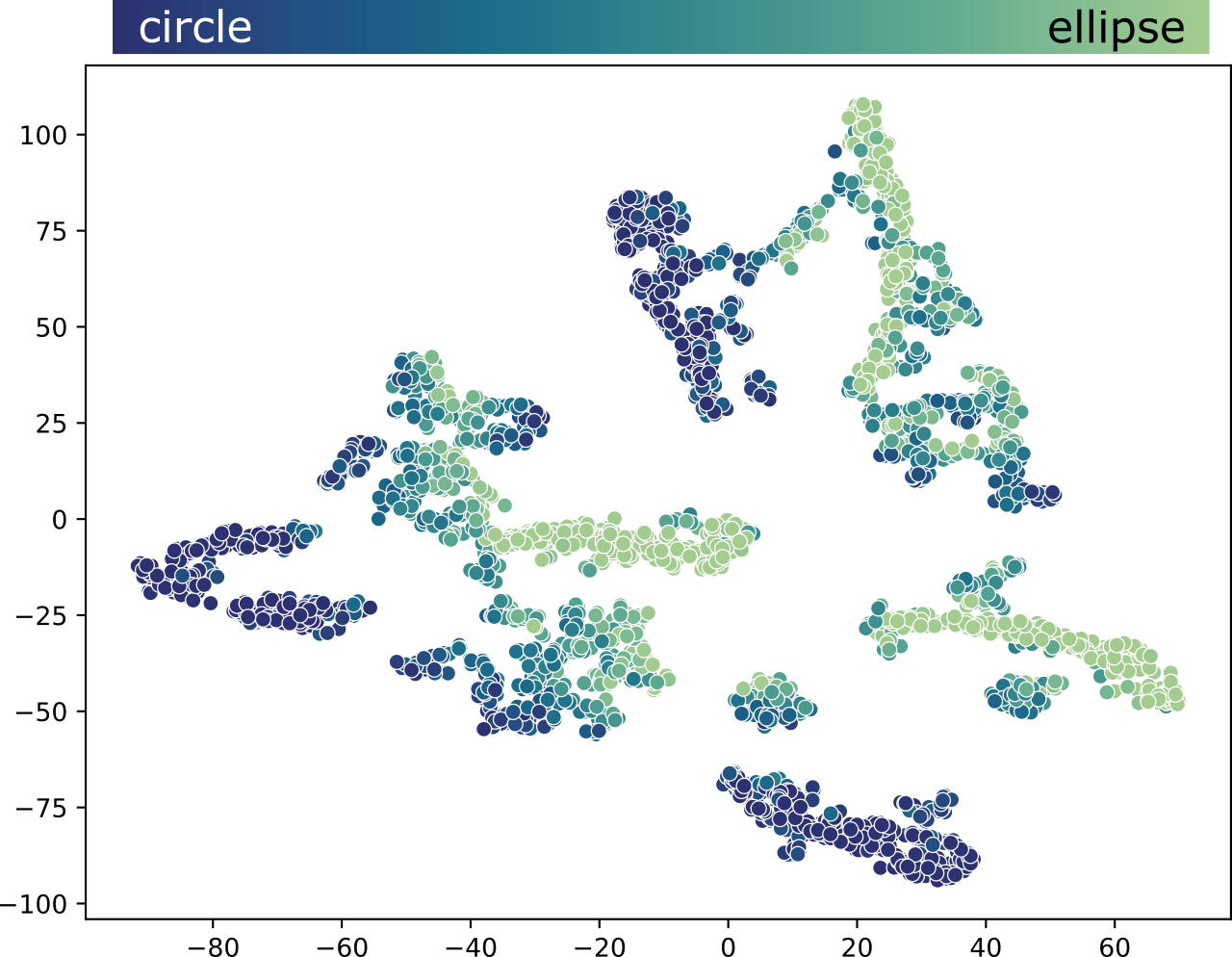}{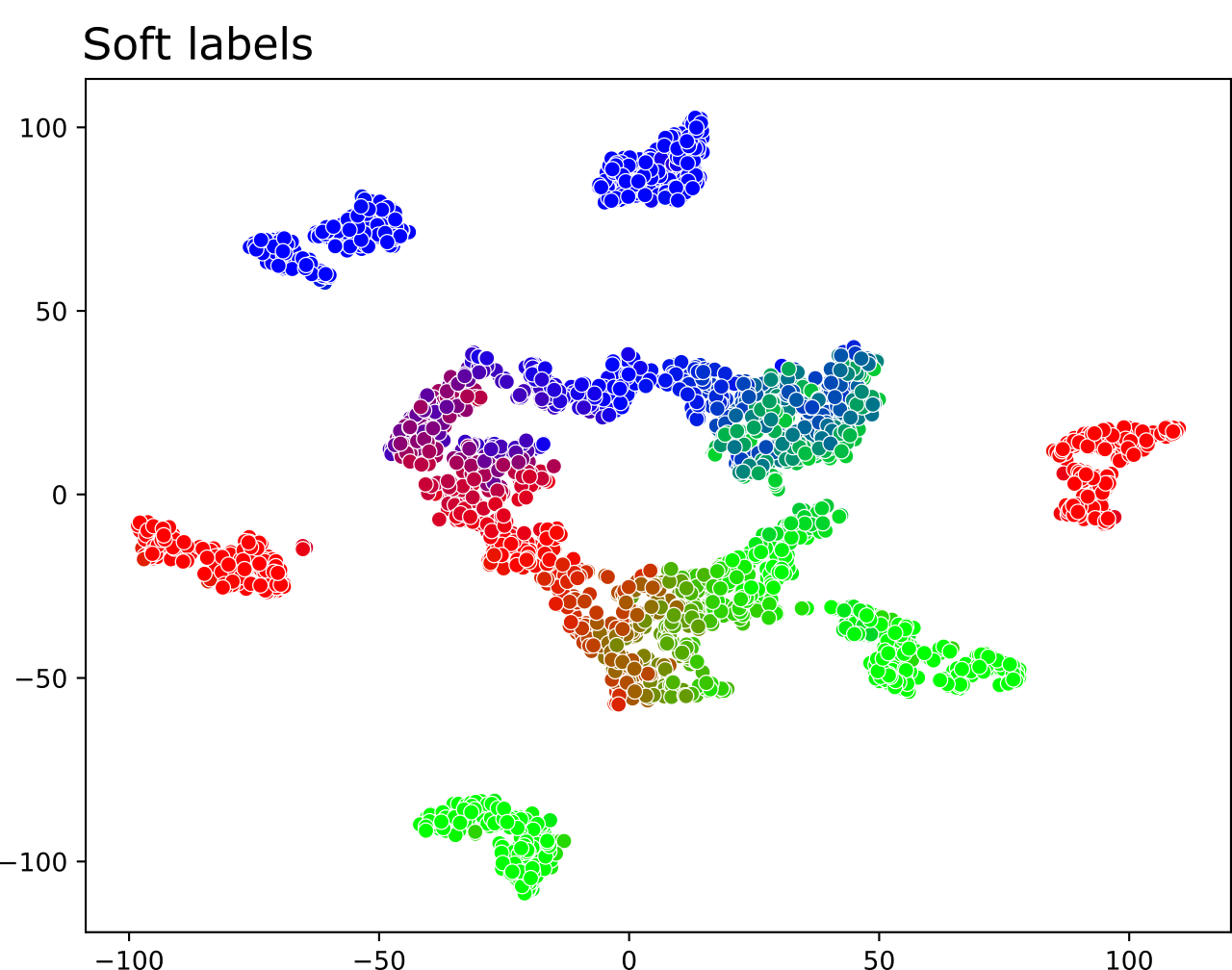}{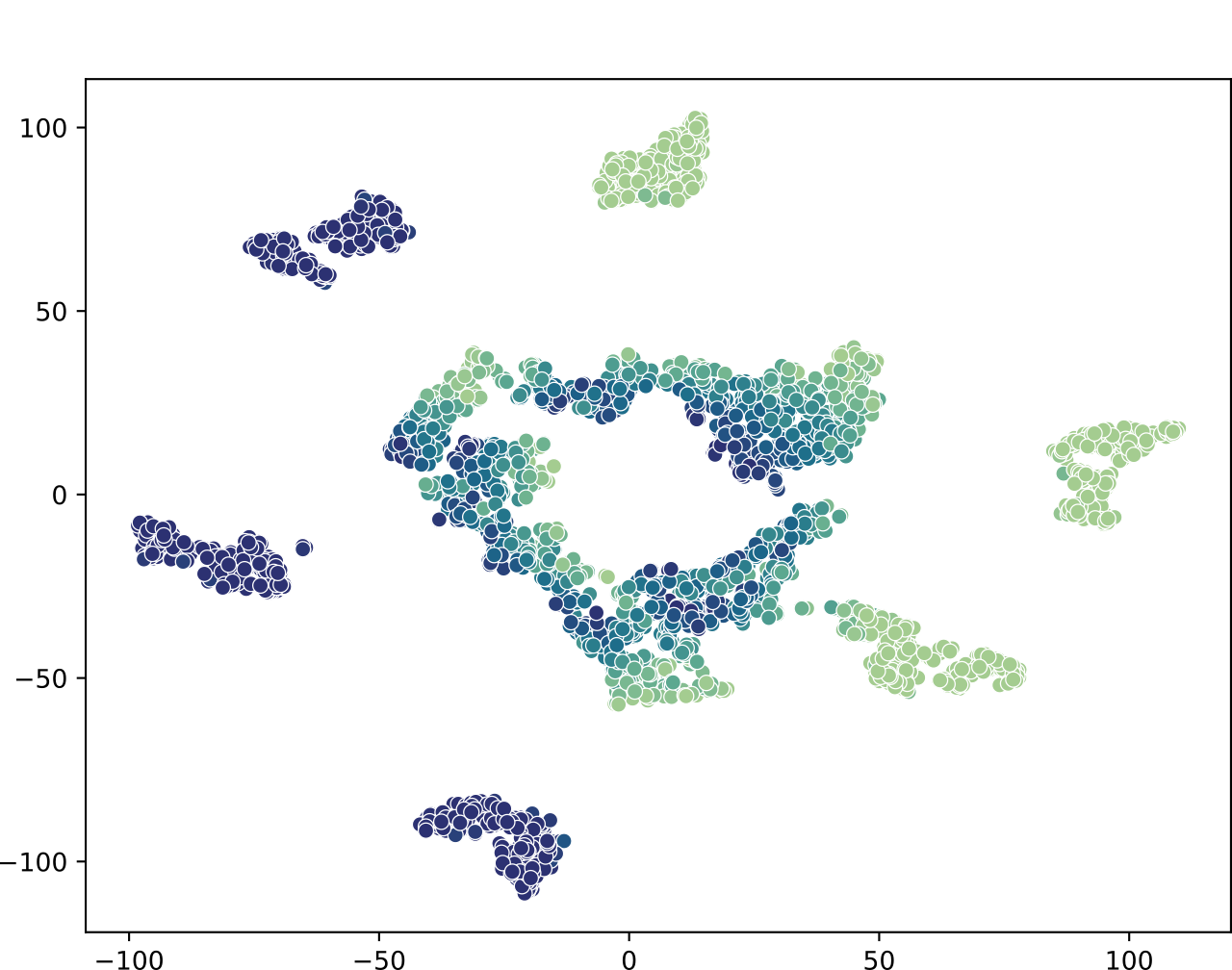}{Using hard labels / color transitions}{Using hard labels / shape transitions}{Using soft labels / color transitions}{Using soft labels / shape transitions}{T-SNE plots for the synthetic dataset using hard / soft labels --
The left images show the ground truth color interpolation, while the right images show the ground truth shape interpolation.
}

Several comparisons on different label types have been published. Tyrväinen showed that models trained on their own soft-labeled CIFAR10 dataset \cite{cifar10h} are more robust against adversarial attacks than models trained on hard-labeled data~\cite{tyrvainen2021soft}. Geng et al. introduced a association classification on soft labels to characterize their imprecisions and lead to more robust classification results~\cite{geng2021arc}. A survey on classifications with different label types is given in~\cite{song2022noisyLabelsSurvey}.

Our main contributions are that we illustrate and discuss (1) the negative impact on data-quality of ambiguous images even without noise  and (2) the insufficient representation of hard labels for ambiguous data.

\legend{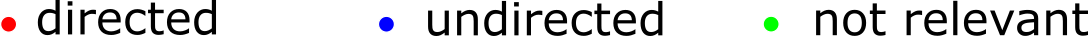}{mice_legend}
\picTwo{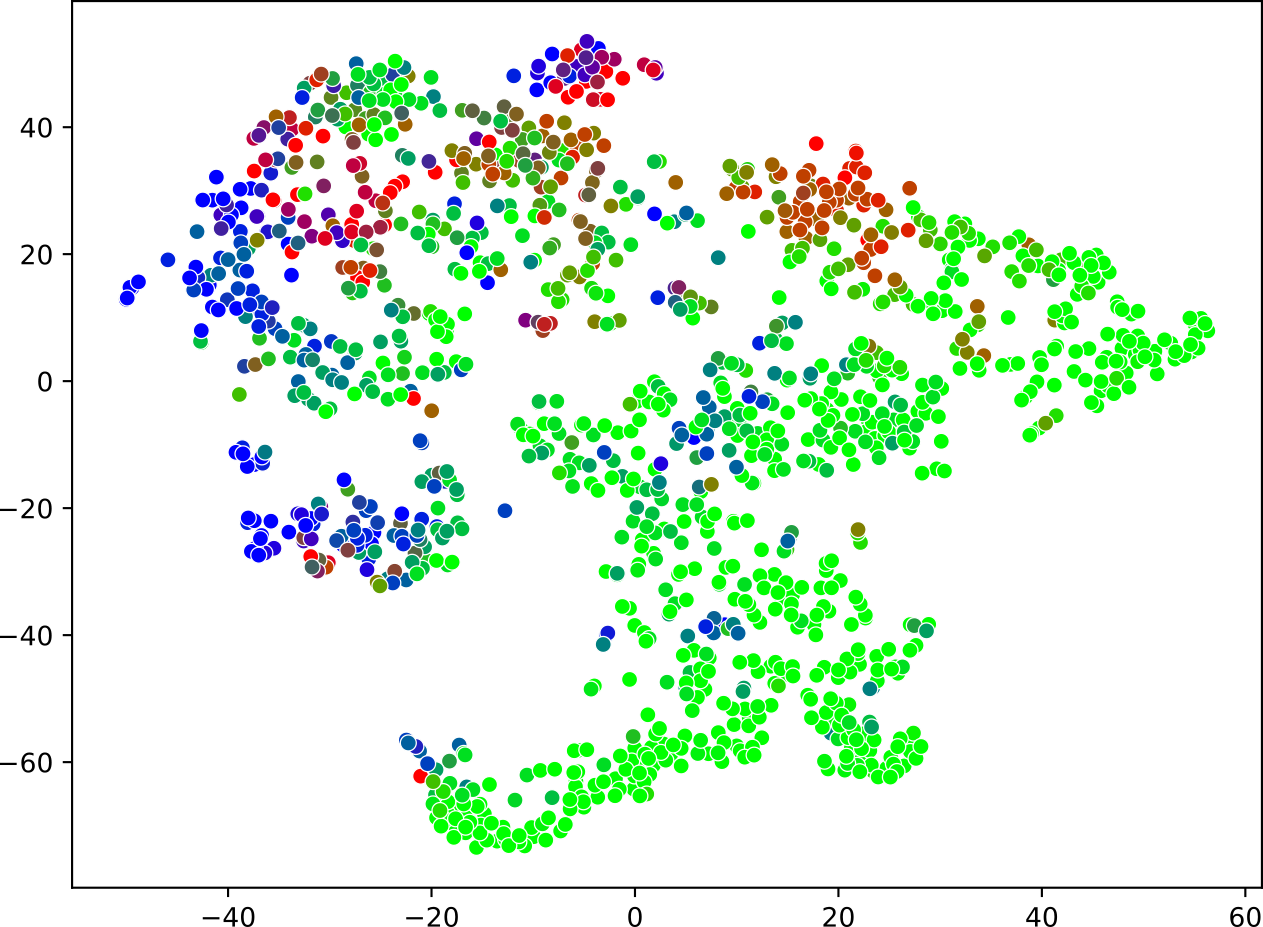}{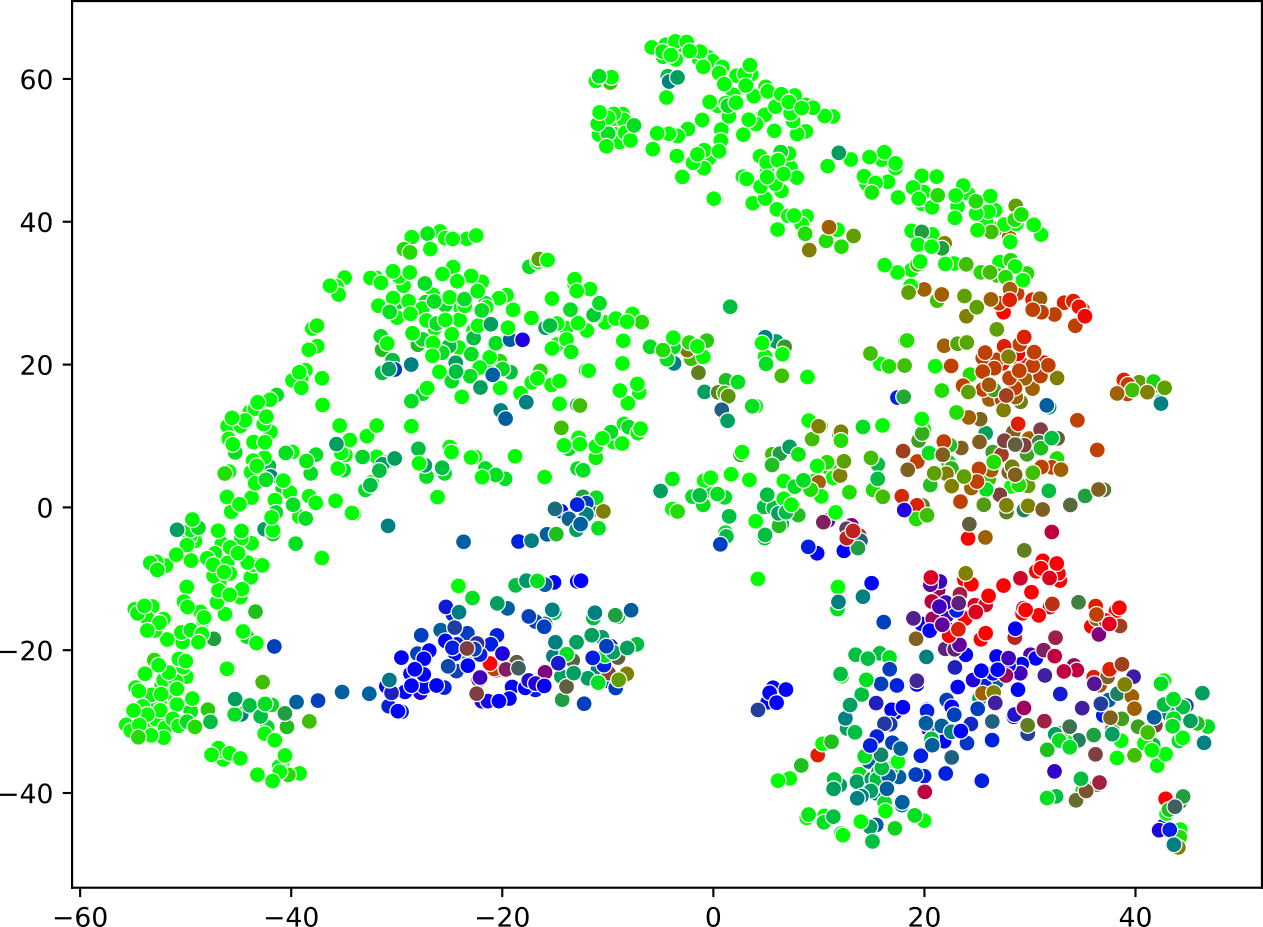}{mice}{Using hard labels}{Using soft labels}{T-SNE plots for the MiceBone dataset using hard / soft labels --
The different colors represent the three classes and the interpolated color directly corresponds to the ground truth class distribution. }

\section{Method}

We compare the effect of hard and soft labels on one synthetic and one real-world dataset for image classification using deep learning.
For a classification problem with $k$ classes and for every image $x$, we use multiple human annotations $a_i(x) \in \{0,1\}^k$ to create the hard and soft label.
The hard label is the (relative) majority vote across all $N$ annotations ($l_h(x) = argmax_k \sum_{i=0}^N a_i(x) \in \{0,...,k\}$) and soft label is the average across all annotations ($l_s(x) = \frac{1}{N} \sum_{i=0}^N a_i(x) \in [0,1]^k$).

As a synthetic dataset, we use red, blue and green circles or ellipses and their color and shape interpolations on a black background.
We generated 15,000 images and used 60\% for training, 20\% for validation and 20\% for testing.
The generated images are either directly one of the six classes (40\%) or an interpolation of the classes (60\%).
Because of our knowledge of the used interpolation, we can estimate the labels perfectly.
This means that we do not have any noise in our data but still have ambiguity.
As a real-world dataset, we used the MiceBone dataset from \cite{dc3} but added more annotations per image for a better label distribution estimation.
We have about 15 annotations on average per image for 7,240 images total and use the same training, validation and test split proportions as for the synthetic dataset.
We have three classes: directed collagen fibers, undirected collagen fibers and not relevant structures.
We have a class imbalance of about 70\% for the non-relevant class and 15\% for the directed and undirected fibers.
We can estimate an image label only based on the collected annotations and therefore have noise and ambiguity in our dataset.

We determined the hyperparameters like model, batch size and learning rate heuristically across the same predefined parameter grid. 
For the synthetic dataset, we used a pretrained DenseNet121 \cite{huang2017densenet} with a batch size of 128, a learning rate of 0.1 for 60 epochs and SGD with cosine learning rates \cite{loshchilov2016cosine} and weight decay of 0.0005.
For the MiceBone dataset, we used a pretrained ResNet50v2 \cite{resnet} with a batch size of 128, a learning rate of 0.1 for 60 epochs, SGD with cosine learning rates, warm restarts \cite{loshchilov2016cosine} and weight decay of 0.001.
All experiments were executed on an RTX 3090 Ti with 24GB VRAM.

We report the macro accuracy (\acc) and Kullback Leiber Divergence (\kl) between the model predictions and the soft ground truth label distributions on the test set.
The macro accuracy is the per class average of true positives divided by the samples of the class.
For the synthetic dataset, this value is equivalent to the accuracy over the complete dataset but is more robust to the class imbalance on the MiceBone dataset.
The Kullback Leiber Divergence is an established metric for measuring the difference between two distributions \cite{murphy2012machinelearning}.
This metric allows a better insight into the difference between the ground truth and predicted distribution, since the accuracy only looks at the most likely class from the distribution.
We report the average and standard deviation across three randomly generated training, validation, and test splits.

For the qualitative comparison, we used the internal global average pooling (GAP) layer features of the trained models. 
We calculated the T-SNE \cite{van2008tsne} plots by aligning their feature distances with a perplexity of 30 over 5,000 iterations. 

\section{Discussion}

\tblResults

We give in \autoref{tbl:datasets} a quantitative comparison for both datasets between using hard and soft labels. 
Across both datasets and metrics, we see that soft labels result in superior results. 
We see an average increase of 7-8\% for \acc, while \kl is reduced by a manifold.
Due to the fact \kl measures the difference in distribution and hard labels collapse the whole training distribution to one label, this difference can be expected. 
However, the increase in \acc indicates that the model can also obtain better majority votes if the input distribution contains the image ambiguity.
We credit this to the possibility of differentiating between images with a high or low ambiguity with the same majority class.
During back propagation with hard labels, both images are enforced to be treated equally, while with soft labels we allow a soft distribution of the error.

Qualitative evaluations using T-SNE plots are given in \autoref{fig:synth_color_hard} and \autoref{fig:mice}.
If we compare the color interpolation in the embedding space for the synthetic dataset, we see smooth transitions for the hard and soft labels.
However, with the soft labels, we see six distinct clusters surrounding a cyclic interpolation region.
This structure exactly represents the data generation process where we generated non-interpolated images and interpolated images.
If we look at the shape interpolations, we see that hard labels generate more individual clusters, while soft labels have a more connected interpolation space.
Only with soft labels we give the model an exact representation of the expected feature space.
With hard labels, the model must create an appropriate feature space based on the majority votes.
These results indicate that the model automatically detects ambiguity in hard labels and tries to shape the feature space accordingly.
Soft labels help structure the feature space.
When we look at the MiceBone results, we can confirm this hypothesis.
For hard labels as input, we see a feature space with clusters of low ambiguity and interpolations for images with high ambiguity. 
However, the separation and transitions are not as clear as in the synthetic dataset. 
When using soft labels as input, we see a better structure and more well-defined transitions between the ambiguous classes.

The found results are of high importance for the data-quality estimation.
On the synthetic dataset, we do not have any noisy and only ambiguity.
Thus, we can credit the performance difference completely to the better representation of soft labels in contrast to hard labels. 
This difference can also be confirmed on a real-world dataset. 
In the qualitative analysis, we showed that soft labels help structure the internal feature space for a better representation of the expected label distribution.
If we look at our motivation, we know that label images suffer from ambiguity and thus it also theoretically impossible to capture this information in just one hard label.
We conclude that soft labels are potentially more suitable representations of label distributions than hard labels.
We should further investigate soft labels and their potential to create higher quality data.
Of special interest is the question, how to obtain such soft labels. 
In many cases it is not feasible to annotate all images multiple times.
Possible solutions for this issue include semi-supervised learning approaches \cite{mean-teacher,fixmatch,simclr} and proposal system \cite{foc,dc3}.

A limitation of our work is that we only used two datasets and standard supervised learning.
We must check whether our conclusions generalize to other datasets and different training protocols.
We also neglected the issue of acquiring the annotations for estimating soft labels.
For many datasets, it is not feasible to create multiple annotations for thousand or even millions of images. 
We are confident that combinations of recent developments  in the field of semi-supervised / self-supervised learning \cite{simclrv2,fixmatch} and soft labels could be used to close the gap of required labeled images to the investigated fully supervised setting.

\bibliography{lib}
\bibliographystyle{icml2022}

\end{document}